\documentclass[sn-mathphys,Numbered]{sn-jnl}%Math and Physical Sciences Reference Style
\usepackage{subcaption}
\usepackage{graphicx}%
\usepackage{multirow}%
\usepackage{pgffor}%
\usepackage{amsmath,amssymb,amsfonts}%
\usepackage{amsthm}%
\usepackage{mathrsfs}%
\usepackage[title]{appendix}%
\usepackage{xcolor}%
\usepackage{textcomp}%
\usepackage{manyfoot}%
\usepackage{booktabs}%
\usepackage{tabularx}%
\usepackage{algorithm}%
\usepackage{adjustbox}
\usepackage{comment}
\usepackage{algorithmicx}%
\usepackage{algpseudocode}%
\usepackage{listings}%
\usepackage{caption}%
\usepackage{array}%
\usepackage{float}%
\theoremstyle{thmstyletwo}%
\theoremstyle{thmstylethree}%
\begin{document}

\title[Article Title]{SurGen: Text-Guided Diffusion Model for Surgical Video Generation}

\author*[1]{\fnm{Cho} \sur{Joseph}}\email{jcho5@stanford.edu}
\author[1, 2]{\fnm{Schmidgall} \sur{Samuel}}
\author[1]{\fnm{Zakka} \sur{Cyril}}
\author[1]{\fnm{Mathur} \sur{Mrudang}}
\author[1]{\fnm{Kaur} \sur{Dhamanpreet}}
\author[2]{\fnm{Shad} \sur{Rohan}}
\author*[1]{\fnm{Hiesinger} \sur{William}}\email{willhies@stanford.edu}

\affil[1]{\orgdiv{Department of Cardiothoracic Surgery}, \orgname{Stanford Medicine}}
\affil[2]{\orgdiv{Department of Electrical \& Computer Engineering}, \orgname{Johns Hopkins University}}
\affil[3]{\orgdiv{Division of Cardiovascular Surgery}, \orgname{Penn Medicine}}

\abstract{Diffusion-based video generation models have made significant strides, producing outputs with improved visual fidelity, temporal coherence, and user control. These advancements hold great promise for improving surgical education by enabling more realistic, diverse, and interactive simulation environments. In this study, we introduce SurGen, a text-guided diffusion model tailored for surgical video synthesis. SurGen produces videos with the highest resolution and longest duration among existing surgical video generation models. We validate the visual and temporal quality of the outputs using standard image and video generation metrics. Additionally, we assess their alignment to the corresponding text prompts through a deep learning classifier trained on surgical data. Our results demonstrate the potential of diffusion models to serve as valuable educational tools for surgical trainees.}
\maketitle
\begin{figure}[H] % The [H] option ensures the figure is placed exactly here.
    \centering
    \includegraphics[width=\linewidth]{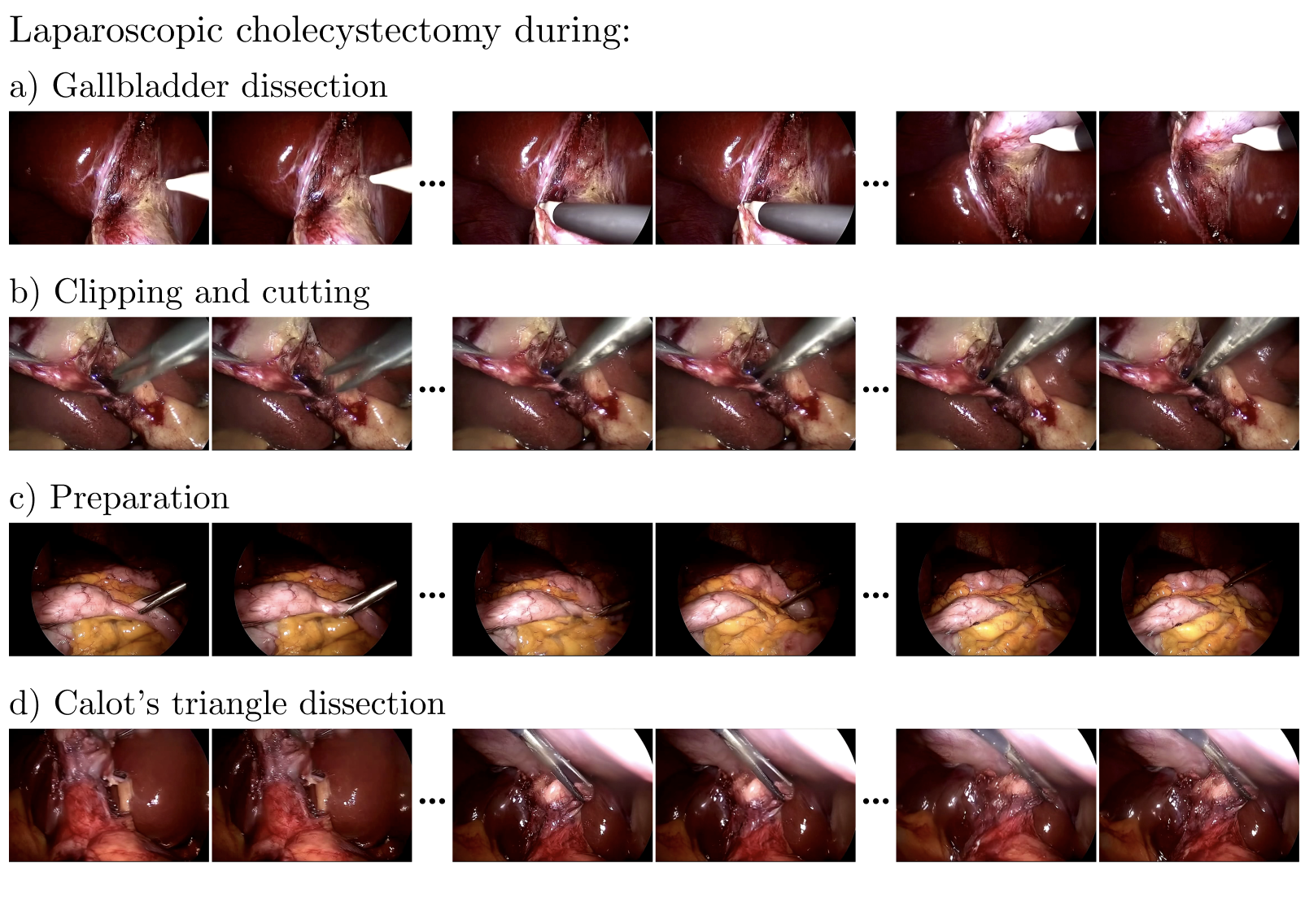} % Adjust the width as needed.
    \caption{Series of videos generated by SurGen, a 2 billion parameter text-guided diffusion model adapted to laparoscopic cholecystectomy procedures. The text prompts for the corresponding videos were formatted as ``Laparoscopic cholecystectomy during \{surgical phase\}". }
    \label{fig:your_label}
\end{figure}

\section{Introduction}\label{sec1}

Generative artificial intelligence (AI) has made significant advancements in healthcare, with applications ranging from radiology report generation \cite{yan2023style,hamamci2024ct2rep} and clinical text summarization \cite{vanveen2024adapted, aali2024benchmark} to text-guided medical image generation \cite{ROENT, cho2024medisyn}. Yet, the field of medical video generation remains largely unexplored. Video synthesis presents challenges that go beyond static image creation, requiring both high visual quality and temporal coherence across frames \cite{li2024survey}. This is particularly important for producing smooth transitions and realistic movements over time.
In surgery, these challenges are compounded by the need for generated videos to accurately represent anatomical structures, deformable tissue dynamics, and surgical instrument movements \cite{Varghese2024}. 

Denoising diffusion probabilistic models (DDPMs) \cite{DDPM}, known for their state-of-the-art performance in image and video generation \cite{dhariwal2021diffusion, bar2024lumiere}, provide a promising approach toward addressing these challenges. These models work by gradually adding random noise to an image or video until it conforms to an isotropic Gaussian distribution. This process is then used to train a neural network to restore the original content from noise. Through iterative denoising, DDPMs are capable of synthesizing novel images and videos from pure noise. In addition, recent progress in text embeddings \cite{CLIP, T5} have enabled DDPMs to integrate textual prompts, offering greater control over the generation process. Although previous works adapted DDPMs for generating high-quality surgical images \cite{allmendinger2023navigating, cho2024medisyn, kaleta2024minimal, nwoye2024surgical}, progress in surgical video generation remains limited. While Li et al. \cite{li2024endora} explored surgical video generation, the synthesized outputs were low resolution (128 x 128 pixels), short duration (16 frames), and lacked input conditioning. Similarly, Iliash et al. \cite{iliash2024interactive} generated surgical videos conditioned on pre-existing instrument masks, which requires real labeled data. 

Here, we propose SurGen, which is the first (to our knowledge) text-guided video generation model for surgery. SurGen leverages a video diffusion architecture based on CogVideoX \cite{yang2024cogvideox} to generate videos of surgical procedures with a higher resolution  (720 x 480 pixels, width x height) and longer duration (49 frames) than existing methods. Our generated videos are conditioned on text prompts that specify distinct surgical phases: preparation, gallbladder dissection, clipping and cutting, and Calot's triangle dissection. We train our model on 200,000 unique frame sequences from laparoscopic cholecystectomy videos, each paired with a phase-specific prompt. We evaluate the visual fidelity and diversity of the model outputs using both the Fr\'echet Inception Distance (FID) \cite{heusel2017ttur} for individual frames and the Fr\'echet Video Distance (FVD) \cite{Unterthiner2018}. Additionally, we compare the phase alignment of the generated videos to the corresponding real data using a 3D Convolutional Neural Network (CNN) trained on a held-out set of laparoscopic cholecystectomy videos.

\section{Related Work}\label{relwork}

Recent progress in image and video generation have been largely driven by the advancement of diffusion models. To overcome initial inefficiencies in pixel-level denoising, latent diffusion models (LDMs) were developed \cite{LDM}, compressing images into a latent space for faster processing. LDMs were extended to video generation \cite{ALIGN} through the addition of temporal processing layers. Recent work has seen transformer architectures replacing traditional U-Nets for denoising operations \cite{peebles2022scalable, Lu2023VDT,esser2024scaling}, resulting in improved image and video quality, greater variation in the generated content, and adherence to text.

Several studies have explored using diffusion models in the surgical domain. For instance, Allmendinger et al. \cite{cho2024medisyn} developed a series of surgical image generation models conditioned on both text and segmentation maps, aiming to use the synthetic images for augmenting existing datasets. Similarly, Nwoye et al. \cite{nwoye2024surgical} adapted Imagen to generate surgical images from triplet action-based textual prompts, also addressing data scarcity concerns. Extending beyond static surgical images, Li et al. adopted an unconditional video diffusion transformer for generating endoscopy videos, including laparoscopic cholecystectomy procedures. Additionally, Iliash et al. used a three-stage method involving Stable Diffusion for surgical text-to-image training, ControlNet for instrument mask conditioning, and ControlVideo for generating surgical videos from both instrument mask frames and text.

Our method improves upon existing work in surgical video generation by 1) conditioning solely on text, thereby eliminating the need for real data inputs, 2) producing higher resolution videos in a wider format, and 3) generating videos of significantly longer duration.

\begin{figure}[H] % The [H] option ensures the figure is placed exactly here.
    \centering
    \includegraphics[width=\linewidth]{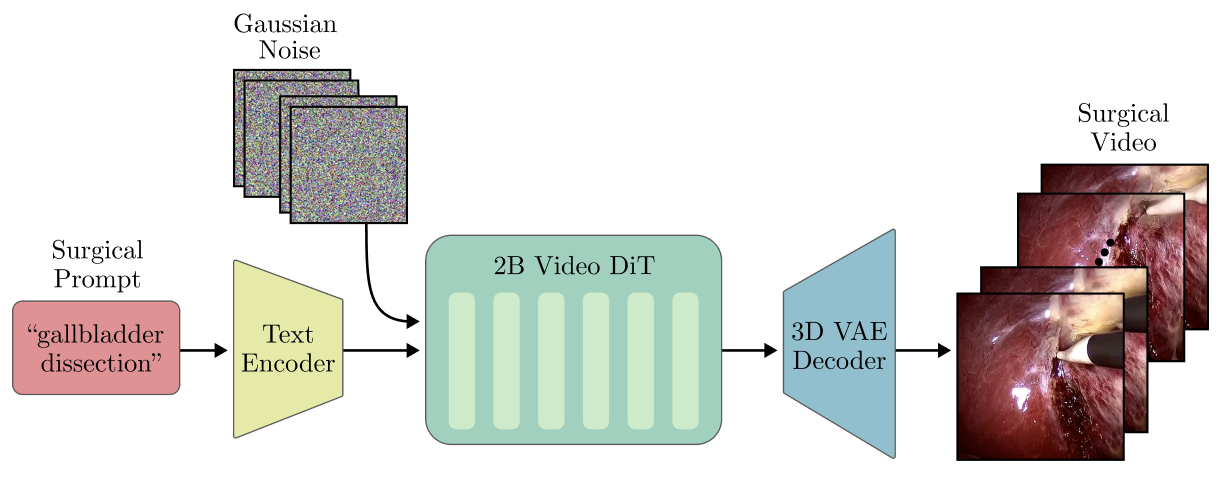} % Adjust the width as needed.
    \caption{The text-to-video process of SurGen, our large video LDM adapted from CogVideoX. The text prompt is processed by a T5 text encoder to create a semantic representation. The diffusion transformer takes in Gaussian noise, and uses the text encoding to help guide the denoising process. The resulting denoised output is then decoded by the 3D VAE into a high-quality surgical video.}
    \label{fig:your_label}
\end{figure}

\section{Methods}\label{methods}
\subsection{Description of the Dataset}
The data used for model training and evaluation is sourced from Cholec80 \cite{twinanda2016endonet}, a dataset of 80 laparoscopic cholecystectomy procedures performed by 13 surgeons. We follow the original train-test split, using the first 40 videos for training and the remaining 40 for evaluation. We extract surgical phase labels—preparation, Calot's triangle dissection, gallbladder dissection, and clipping and cutting—to create 200,000 video-text pairs for training. Specifically, for each surgical phase, we extract 50,000 unique sequences consisting of 49 frames each, with each frame in the sequence spaced two frames apart from the original video.

\subsection{Data Pre-processing}
In all video sequences, we crop each frame from its original width of 840 pixels down to 720 pixels, while maintaining its original height at 480 pixels. This effectively removes much of the black borders typical of endoscopic footage, ensuring that all essential surgical details are retained. We format the corresponding text prompts as ``Laparoscopic cholecystectomy during \{surgical phase\}".

\begin{figure}[H] % The [H] option ensures the figure is placed exactly here.
    \centering
    \includegraphics[width=\linewidth]{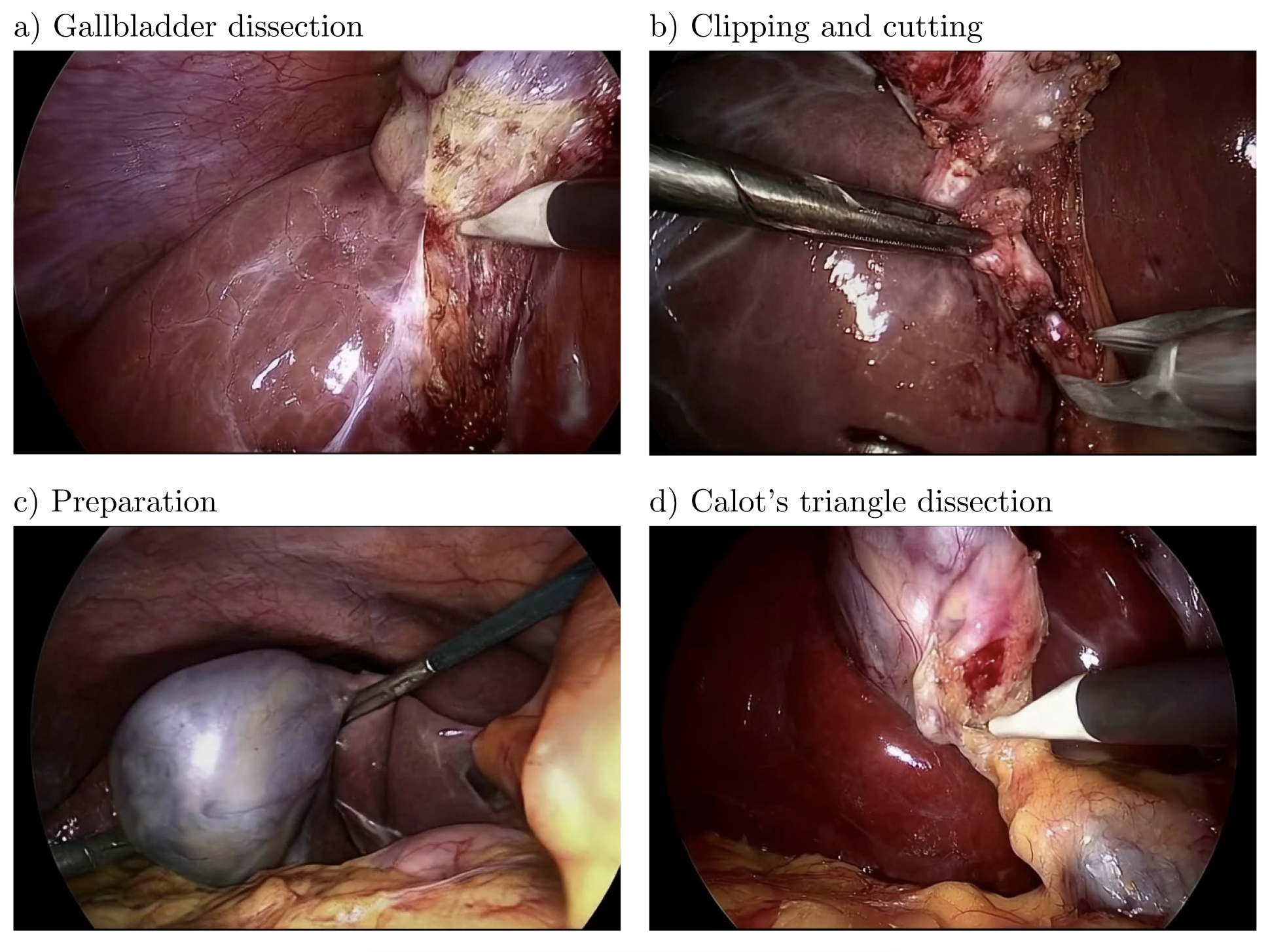} % Adjust the width as needed.
    \caption{High-resolution frames from the surgical videos synthesized by SurGen. The model generates videos at 720 x 480 pixels (width × height).}
    \label{fig:your_label}
\end{figure}

\subsection{Model Architecture and Training}
For our video generation model, we adopt CogVideoX, a 2 billion parameter text-guided LDM. CogVideoX combines three major components to synthesize a video conditioned on an text prompt:  
\begin{itemize}
    \item \textbf{3D Variational Autoencoder}: To accelerate denoising operations, the encoder of a 3D Variational Autoencoder (VAE) \cite{kingma2013auto, yu2023language} compresses each video into a latent space, reducing its spatial dimensions by 8x and temporal dimensions by 4x. The decoder of the 3D VAE converts the denoised representations into full video frames. 
    
    \item \textbf{Denoising Video Transformer}: A 2 billion parameter text-conditional video transformer \cite{arnab2021vivit} is used to denoise the latent vectors. Notably, the model uses a full 3D attention mechanism, allowing spatial-temporal patches to attend to each other across all locations.
    
 \item \textbf{Text Encoder}: A T5 text encoder \cite{raffel2019exploring} converts the text prompts into semantically rich representations, which are then fed into the diffusion transformer to guide the denosing process.

\end{itemize}

In training our model, we initialize all its components with pretrained weights from the publicly released base model in an effort to reduce training time. As with other approaches in developing text-guided medical image and video generation models \cite{Chambon2022AdaptingPV, cho2024medisyn, abaid2024synthesizing}, we keep the text encoder and VAE frozen, training only the denoising model. We fine-tune the video transformer for 50,000 steps with an effective batch size of 4, minimizing the Mean Squared Error (MSE) Loss between its predicted noise and the corresponding ground truth. The model is trained using a fused AdamW \cite{loshchilov2017decoupled} optimizer with a learning rate of 2e-4, a weight decay of 1e-4, and an epsilon value of 1e-8. Distributed data training is performed across 4 A100 GPUs, spanning roughly 3 days.

\maketitle
\begin{figure}[H] % The [H] option ensures the figure is placed exactly here.
    \centering
    \includegraphics[width=\linewidth]{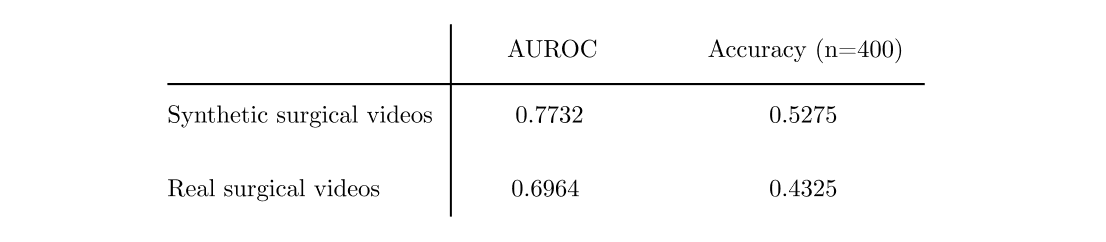} % Adjust the width as needed.
    \caption{Evaluation results of a 3D ResNet18, trained on the last 40 videos of Cholec80, show higher top-1 accuracy and AUROC in classifying surgical phases for SurGen-synthesized videos compared to the first 40 videos of Cholec80 (used to train SurGen). }
    \label{fig:your_label}
\end{figure}

\subsection{Evaluation}

To evaluate the generated videos, we first use standard image and video generation metrics, including the FID on individual frames and FVD. FID evaluates the visual quality and diversity of the generated images relative to the original dataset, while the FVD extends this metric by also accounting for differences in temporal dynamics, enabling a direct comparison of videos. For the analysis, we generate 2048 videos, evenly distributed across the four surgical phases in the training dataset. We also extract another set of 2048 sequences, maintaining the same distribution of phases, from the evaluation set of Cholec80 to serve as our comparison dataset. We calculate the FID on the middle frames of the sequences and the FVD on the first 16 frames of the videos. As there are no surgical video generation models trained on Cholec80, we use the base CogVideoX model as our baseline for comparison.

In assessing the phase alignment of the synthesized videos, we use a 3D ResNet18 \cite{he2016deep, hara2017learning} model previously trained on the Kinetics dataset \cite{kay2017kinetics}. We fine-tune this model for 15 epochs on 2,000 sequences of 16 frames from the last 40 videos of Cholec80, with the videos evenly distributed across the four surgical phases. We then calculate and compare the model's top-1 accuracy and area under the Receiver Operating Characteristic curve (AUROC) on 400 generated videos and 400 videos randomly sampled from SurGen's training dataset, both with an even phase distribution.

\section{Results}
The SurGen model achieves a substantially lower FID of 79.9163 compared to its pre-trained version's score of 260.0301, indicating improved visual quality and diversity. Additionally, SurGen obtains a significantly lower FVD of 752.7587 versus 1975.5990 for the pre-trained model, suggesting SurGen's generated surgical videos more closely resembles the temporal dynamics of real surgical footage. 

In evaluating the surgical phase-alignment of the generated videos compared to their real counterparts, we found that the fine-tuned 3D ResNet18 model actually performed better on the generated videos in terms of both accuracy (0.5275 vs. 0.4325) and AUROC (0.7732 vs. 0.6964). We suspect that the generated videos exhibited stronger alignment to their corresponding phases because the diffusion model effectively captured the common visual and temporal patterns of each phase, potentially resulting in fewer outliers generated.

\section{Discussion} 

In this work, we introduce SurGen, the first (to our knowledge) text-guided diffusion model for generating surgical videos. Improvements in standard image and video generation metrics suggest that the model effectively captures both the visual and temporal characteristics, as well as the overall variety in the surgical videos. Additionally, the generated outputs' strong alignment with their corresponding surgical phases highlights the model's ability to differentiate and accurately represent complex surgical concepts. By synthesizing high-resolution videos at a longer duration, SurGen demonstrates the potential of diffusion-based video generation models to synthesize surgical footage without compromising the original quality.

Previous work has shown that generative models can be used as simulators for generating action-guided data \cite{kim2021drivegan, hu2023gaia, wang2023drivedreamer}. Currently, existing virtual reality simulators for surgical procedures are hand-designed and are therefore limited to predefined scenarios with limited physical realism \cite{Badash2016}. By leveraging video generation models, we can potentially simulate highly realistic surgical procedures that display a wide range of disease severity and anatomical variations. Furthermore, with improved control through text prompts, surgical trainees could select and customize the types of scenarios they wish to practice in. This would better equip them to handle a diverse array of surgical cases, thereby improving postoperative outcomes for patients.

Our study has several limitations. First, due to the scarcity of labeled surgical data, our model was trained on only 40 laparoscopic cholecystectomy videos. To more accurately represent the physical dynamics in surgical procedures, extensive training on a larger and more diverse surgical dataset is required. Second, the model's input conditioning was limited to surgical phases. To add more customization and applicability, factors such as disease severity and patient demographics should be incorporated into the text prompts, though this would require additional data labeling. Third, our model currently lacks kinematic conditioning (specifically instrument movements), which is necessary for enabling responsive interaction with surgeon movements. Last, our model does not generate videos quickly enough for real-time applications. Achieving real-time video generation is essential for creating an immersive surgical training environment.

Future work should aim to increase both the scale of the data and the architectures used for higher quality surgical video generation. Additionally, efforts should focus on developing generative models that dynamically respond to kinematic inputs. We believe these advancements are necessary for creating a highly realistic, diverse, and responsive simulation environment for surgical trainees. 

\section{Acknowledgments}
We would like to thank Stanford Sherlock for their continuous support with GPU access.
\section*{Declarations}
\subsection{Funding}
This project was supported in part by a National Heart, Lung, and Blood Institute (NIH NHLBI) grant (1R01HL157235-01A1) (W.H.).

\subsection{Competing interests}
The authors declare no competing interests.

\bibliography{sn-bibliography}% common bib file


\begin{thebibliography}{41}
\ifx \bisbn   \undefined \def \bisbn  #1{ISBN #1}\fi
\ifx \binits  \undefined \def \binits#1{#1}\fi
\ifx \bauthor  \undefined \def \bauthor#1{#1}\fi
\ifx \batitle  \undefined \def \batitle#1{#1}\fi
\ifx \bjtitle  \undefined \def \bjtitle#1{#1}\fi
\ifx \bvolume  \undefined \def \bvolume#1{\textbf{#1}}\fi
\ifx \byear  \undefined \def \byear#1{#1}\fi
\ifx \bissue  \undefined \def \bissue#1{#1}\fi
\ifx \bfpage  \undefined \def \bfpage#1{#1}\fi
\ifx \blpage  \undefined \def \blpage #1{#1}\fi
\ifx \burl  \undefined \def \burl#1{\textsf{#1}}\fi
\ifx \doiurl  \undefined \def \doiurl#1{\url{https://doi.org/#1}}\fi
\ifx \betal  \undefined \def \betal{\textit{et al.}}\fi
\ifx \binstitute  \undefined \def \binstitute#1{#1}\fi
\ifx \binstitutionaled  \undefined \def \binstitutionaled#1{#1}\fi
\ifx \bctitle  \undefined \def \bctitle#1{#1}\fi
\ifx \beditor  \undefined \def \beditor#1{#1}\fi
\ifx \bpublisher  \undefined \def \bpublisher#1{#1}\fi
\ifx \bbtitle  \undefined \def \bbtitle#1{#1}\fi
\ifx \bedition  \undefined \def \bedition#1{#1}\fi
\ifx \bseriesno  \undefined \def \bseriesno#1{#1}\fi
\ifx \blocation  \undefined \def \blocation#1{#1}\fi
\ifx \bsertitle  \undefined \def \bsertitle#1{#1}\fi
\ifx \bsnm \undefined \def \bsnm#1{#1}\fi
\ifx \bsuffix \undefined \def \bsuffix#1{#1}\fi
\ifx \bparticle \undefined \def \bparticle#1{#1}\fi
\ifx \barticle \undefined \def \barticle#1{#1}\fi
\bibcommenthead
\ifx \bconfdate \undefined \def \bconfdate #1{#1}\fi
\ifx \botherref \undefined \def \botherref #1{#1}\fi
\ifx \url \undefined \def \url#1{\textsf{#1}}\fi
\ifx \bchapter \undefined \def \bchapter#1{#1}\fi
\ifx \bbook \undefined \def \bbook#1{#1}\fi
\ifx \bcomment \undefined \def \bcomment#1{#1}\fi
\ifx \oauthor \undefined \def \oauthor#1{#1}\fi
\ifx \citeauthoryear \undefined \def \citeauthoryear#1{#1}\fi
\ifx \endbibitem  \undefined \def \endbibitem {}\fi
\ifx \bconflocation  \undefined \def \bconflocation#1{#1}\fi
\ifx \arxivurl  \undefined \def \arxivurl#1{\textsf{#1}}\fi
\csname PreBibitemsHook\endcsname

%%% 1
\bibitem[\protect\citeauthoryear{Yan et~al.}{2023}]{yan2023style}
\begin{barticle}
\bauthor{\bsnm{Yan}, \binits{B.}},
\bauthor{\bsnm{Liu}, \binits{R.}},
\bauthor{\bsnm{Kuo}, \binits{D.E.}},
\bauthor{\bsnm{Adithan}, \binits{S.}},
\bauthor{\bsnm{Reis}, \binits{E.P.}},
\bauthor{\bsnm{Kwak}, \binits{S.}},
\bauthor{\bsnm{Venugopal}, \binits{V.K.}},
\bauthor{\bsnm{O'Connell}, \binits{C.P.}},
\bauthor{\bsnm{Saenz}, \binits{A.}},
\bauthor{\bsnm{Rajpurkar}, \binits{P.}},
\bauthor{\bsnm{Moor}, \binits{M.}}:
\batitle{Style-aware radiology report generation with radgraph and few-shot prompting}.
\bjtitle{arXiv preprint arXiv:2310.17811}
(\byear{2023})
\doiurl{10.48550/arXiv.2310.17811} .
\bcomment{Accepted to Findings of EMNLP 2023}
\end{barticle}
\endbibitem

%%% 2
\bibitem[\protect\citeauthoryear{Hamamci et~al.}{2024}]{hamamci2024ct2rep}
\begin{botherref}
\oauthor{\bsnm{Hamamci}, \binits{I.E.}},
\oauthor{\bsnm{Er}, \binits{S.}},
\oauthor{\bsnm{Menze}, \binits{B.}}:
Ct2rep: Automated radiology report generation for 3d medical imaging.
arXiv preprint arXiv:2403.06801
(2024).
arXiv:2403.06801 [eess.IV]
\end{botherref}
\endbibitem

%%% 3
\bibitem[\protect\citeauthoryear{Veen et~al.}{2024}]{vanveen2024adapted}
\begin{barticle}
\bauthor{\bsnm{Veen}, \binits{D.V.}},
\bauthor{\bsnm{Uden}, \binits{C.V.}},
\bauthor{\bsnm{Blankemeier}, \binits{L.}},
\bauthor{\bsnm{Delbrouck}, \binits{J.-B.}},
\bauthor{\bsnm{Aali}, \binits{A.}},
\bauthor{\bsnm{Bluethgen}, \binits{C.}},
\bauthor{\bsnm{Pareek}, \binits{A.}},
\bauthor{\bsnm{Polacin}, \binits{M.}},
\bauthor{\bsnm{Reis}, \binits{E.P.}},
\bauthor{\bsnm{Seehofnerova}, \binits{A.}},
\bauthor{\bsnm{Rohatgi}, \binits{N.}},
\bauthor{\bsnm{Hosamani}, \binits{P.}},
\bauthor{\bsnm{Collins}, \binits{W.}},
\bauthor{\bsnm{Ahuja}, \binits{N.}},
\bauthor{\bsnm{Langlotz}, \binits{C.P.}},
\bauthor{\bsnm{Hom}, \binits{J.}},
\bauthor{\bsnm{Gatidis}, \binits{S.}},
\bauthor{\bsnm{Pauly}, \binits{J.}},
\bauthor{\bsnm{Chaudhari}, \binits{A.S.}}:
\batitle{Adapted large language models can outperform medical experts in clinical text summarization}.
\bjtitle{arXiv preprint arXiv:2309.07430}
(\byear{2024})
\doiurl{10.1038/s41591-024-02855-5} .
\bcomment{arXiv:2309.07430 [cs.CL]}
\end{barticle}
\endbibitem

%%% 4
\bibitem[\protect\citeauthoryear{Aali et~al.}{2024}]{aali2024benchmark}
\begin{botherref}
\oauthor{\bsnm{Aali}, \binits{A.}},
\oauthor{\bsnm{Veen}, \binits{D.V.}},
\oauthor{\bsnm{Arefeen}, \binits{Y.I.}},
\oauthor{\bsnm{Hom}, \binits{J.}},
\oauthor{\bsnm{Bluethgen}, \binits{C.}},
\oauthor{\bsnm{Reis}, \binits{E.P.}},
\oauthor{\bsnm{Gatidis}, \binits{S.}},
\oauthor{\bsnm{Clifford}, \binits{N.}},
\oauthor{\bsnm{Daws}, \binits{J.}},
\oauthor{\bsnm{Tehrani}, \binits{A.S.}},
\oauthor{\bsnm{Kim}, \binits{J.}},
\oauthor{\bsnm{Chaudhari}, \binits{A.S.}}:
A benchmark of domain-adapted large language models for generating brief hospital course summaries.
arXiv preprint arXiv:2403.05720
(2024).
arXiv:2403.05720 [cs.CL]
\end{botherref}
\endbibitem

%%% 5
\bibitem[\protect\citeauthoryear{Chambon et~al.}{2022}]{ROENT}
\begin{botherref}
\oauthor{\bsnm{Chambon}, \binits{P.}},
\oauthor{\bsnm{Bluethgen}, \binits{C.}},
\oauthor{\bsnm{Delbrouck}, \binits{J.-B.}},
\oauthor{\bsnm{Sluijs}, \binits{R.V.}},
\oauthor{\bsnm{Połacin}, \binits{M.}},
\oauthor{\bsnm{Chaves}, \binits{J.M.Z.}},
\oauthor{\bsnm{Abraham}, \binits{T.M.}},
\oauthor{\bsnm{Purohit}, \binits{S.}},
\oauthor{\bsnm{Langlotz}, \binits{C.P.}},
\oauthor{\bsnm{Chaudhari}, \binits{A.}}:
Roentgen: Vision-language foundation model for chest x-ray generation
(2022)
{\href{https://arxiv.org/abs/2211.12737}{{arXiv:2211.12737}}}
{[cs.CV]}
\end{botherref}
\endbibitem

%%% 6
\bibitem[\protect\citeauthoryear{Cho et~al.}{2024}]{cho2024medisyn}
\begin{botherref}
\oauthor{\bsnm{Cho}, \binits{J.}},
\oauthor{\bsnm{Zakka}, \binits{C.}},
\oauthor{\bsnm{Kaur}, \binits{D.}},
\oauthor{\bsnm{Shad}, \binits{R.}},
\oauthor{\bsnm{Wightman}, \binits{R.}},
\oauthor{\bsnm{Chaudhari}, \binits{A.}},
\oauthor{\bsnm{Hiesinger}, \binits{W.}}:
Medisyn: Text-guided diffusion models for broad medical 2d and 3d image synthesis
(2024)
{\href{https://arxiv.org/abs/2405.09806}{{arXiv:2405.09806}}}
{[cs.CV]}
\end{botherref}
\endbibitem

%%% 7
\bibitem[\protect\citeauthoryear{Li et~al.}{2024}]{li2024survey}
\begin{botherref}
\oauthor{\bsnm{Li}, \binits{C.}},
\oauthor{\bsnm{Huang}, \binits{D.}},
\oauthor{\bsnm{Lu}, \binits{Z.}},
\oauthor{\bsnm{Xiao}, \binits{Y.}},
\oauthor{\bsnm{Pei}, \binits{Q.}},
\oauthor{\bsnm{Bai}, \binits{L.}}:
A survey on long video generation: Challenges, methods, and prospects.
arXiv preprint arXiv:2403.16407
(2024).
arXiv:2403.16407 [cs.CV]
\end{botherref}
\endbibitem

%%% 8
\bibitem[\protect\citeauthoryear{Varghese et~al.}{2024}]{Varghese2024}
\begin{barticle}
\bauthor{\bsnm{Varghese}, \binits{C.}},
\bauthor{\bsnm{Harrison}, \binits{E.M.}},
\bauthor{\bsnm{O'Grady}, \binits{G.}}, \betal:
\batitle{Artificial intelligence in surgery}.
\bjtitle{Nature Medicine}
\bvolume{30},
\bfpage{1257}--\blpage{1268}
(\byear{2024})
\doiurl{10.1038/s41591-024-02970-3} .
\bcomment{Received 24 January 2024; Accepted 03 April 2024; Published 13 May 2024}
\end{barticle}
\endbibitem

%%% 9
\bibitem[\protect\citeauthoryear{Ho et~al.}{2020}]{DDPM}
\begin{botherref}
\oauthor{\bsnm{Ho}, \binits{J.}},
\oauthor{\bsnm{Jain}, \binits{A.}},
\oauthor{\bsnm{Abbeel}, \binits{P.}}:
Denoising diffusion probabilistic models
(2020)
{\href{https://arxiv.org/abs/2006.11239}{{arXiv:2006.11239}}}
{[cs.LG]}
\end{botherref}
\endbibitem

%%% 10
\bibitem[\protect\citeauthoryear{Dhariwal and Nichol}{2021}]{dhariwal2021diffusion}
\begin{botherref}
\oauthor{\bsnm{Dhariwal}, \binits{P.}},
\oauthor{\bsnm{Nichol}, \binits{A.}}:
Diffusion models beat gans on image synthesis.
arXiv preprint arXiv:2105.05233
(2021).
arXiv:2105.05233 [cs.LG]
\end{botherref}
\endbibitem

%%% 11
\bibitem[\protect\citeauthoryear{Bar-Tal et~al.}{2024}]{bar2024lumiere}
\begin{botherref}
\oauthor{\bsnm{Bar-Tal}, \binits{O.}},
\oauthor{\bsnm{Chefer}, \binits{H.}},
\oauthor{\bsnm{Tov}, \binits{O.}},
\oauthor{\bsnm{Herrmann}, \binits{C.}},
\oauthor{\bsnm{Paiss}, \binits{R.}},
\oauthor{\bsnm{Zada}, \binits{S.}},
\oauthor{\bsnm{Ephrat}, \binits{A.}},
\oauthor{\bsnm{Hur}, \binits{J.}},
\oauthor{\bsnm{Liu}, \binits{G.}},
\oauthor{\bsnm{Raj}, \binits{A.}},
\oauthor{\bsnm{Li}, \binits{Y.}},
\oauthor{\bsnm{Rubinstein}, \binits{M.}},
\oauthor{\bsnm{Michaeli}, \binits{T.}},
\oauthor{\bsnm{Wang}, \binits{O.}},
\oauthor{\bsnm{Sun}, \binits{D.}},
\oauthor{\bsnm{Dekel}, \binits{T.}},
\oauthor{\bsnm{Mosseri}, \binits{I.}}:
Lumiere: A space-time diffusion model for video generation.
arXiv preprint arXiv:2401.12945
(2024).
arXiv:2401.12945 [cs.CV]
\end{botherref}
\endbibitem

%%% 12
\bibitem[\protect\citeauthoryear{Radford et~al.}{2021}]{CLIP}
\begin{botherref}
\oauthor{\bsnm{Radford}, \binits{A.}},
\oauthor{\bsnm{Kim}, \binits{J.W.}},
\oauthor{\bsnm{Hallacy}, \binits{C.}},
\oauthor{\bsnm{Ramesh}, \binits{A.}},
\oauthor{\bsnm{Goh}, \binits{G.}},
\oauthor{\bsnm{Agarwal}, \binits{S.}},
\oauthor{\bsnm{Sastry}, \binits{G.}},
\oauthor{\bsnm{Askell}, \binits{A.}},
\oauthor{\bsnm{Mishkin}, \binits{P.}},
\oauthor{\bsnm{Clark}, \binits{J.}},
\oauthor{\bsnm{Krueger}, \binits{G.}},
\oauthor{\bsnm{Sutskever}, \binits{I.}}:
Learning transferable visual models from natural language supervision
(2021)
\doiurl{10.48550/arXiv.2103.00020}
{[cs.CV]}
\end{botherref}
\endbibitem

%%% 13
\bibitem[\protect\citeauthoryear{Raffel et~al.}{2019}]{T5}
\begin{botherref}
\oauthor{\bsnm{Raffel}, \binits{C.}},
\oauthor{\bsnm{Shazeer}, \binits{N.}},
\oauthor{\bsnm{Roberts}, \binits{A.}},
\oauthor{\bsnm{Lee}, \binits{K.}},
\oauthor{\bsnm{Narang}, \binits{S.}},
\oauthor{\bsnm{Matena}, \binits{M.}},
\oauthor{\bsnm{Zhou}, \binits{Y.}},
\oauthor{\bsnm{Li}, \binits{W.}},
\oauthor{\bsnm{Liu}, \binits{P.J.}}:
Exploring the limits of transfer learning with a unified text-to-text transformer
(2019)
\doiurl{10.48550/arXiv.1910.10683}
{[cs.LG]}
\end{botherref}
\endbibitem

%%% 14
\bibitem[\protect\citeauthoryear{Allmendinger et~al.}{2023}]{allmendinger2023navigating}
\begin{botherref}
\oauthor{\bsnm{Allmendinger}, \binits{S.}},
\oauthor{\bsnm{Hemmer}, \binits{P.}},
\oauthor{\bsnm{Queisner}, \binits{M.}},
\oauthor{\bsnm{Sauer}, \binits{I.}},
\oauthor{\bsnm{Müller}, \binits{L.}},
\oauthor{\bsnm{Jakubik}, \binits{J.}},
\oauthor{\bsnm{V{\"o}ssing}, \binits{M.}},
\oauthor{\bsnm{K{\"u}hl}, \binits{N.}}:
Navigating the synthetic realm: Harnessing diffusion-based models for laparoscopic text-to-image generation.
arXiv preprint arXiv:2312.03043
(2023).
arXiv:2312.03043 [eess.IV]
\end{botherref}
\endbibitem

%%% 15
\bibitem[\protect\citeauthoryear{Kaleta et~al.}{2024}]{kaleta2024minimal}
\begin{barticle}
\bauthor{\bsnm{Kaleta}, \binits{J.}},
\bauthor{\bsnm{Dall'Alba}, \binits{D.}},
\bauthor{\bsnm{Płotka}, \binits{S.}},
\bauthor{\bsnm{Korzeniowski}, \binits{P.}}:
\batitle{Minimal data requirement for realistic endoscopic image generation with stable diffusion}.
\bjtitle{International Journal of Computer Assisted Radiology and Surgery}
\bvolume{19}(\bissue{3}),
\bfpage{531}--\blpage{539}
(\byear{2024})
\doiurl{10.1007/s11548-023-03030-w} .
\bcomment{Epub 2023 Nov 7}
\end{barticle}
\endbibitem

%%% 16
\bibitem[\protect\citeauthoryear{Nwoye et~al.}{2024}]{nwoye2024surgical}
\begin{botherref}
\oauthor{\bsnm{Nwoye}, \binits{C.I.}},
\oauthor{\bsnm{Bose}, \binits{R.}},
\oauthor{\bsnm{Elgohary}, \binits{K.}},
\oauthor{\bsnm{Arboit}, \binits{L.}},
\oauthor{\bsnm{Carlino}, \binits{G.}},
\oauthor{\bsnm{Lavanchy}, \binits{J.L.}},
\oauthor{\bsnm{Mascagni}, \binits{P.}},
\oauthor{\bsnm{Padoy}, \binits{N.}}:
Surgical text-to-image generation.
arXiv preprint arXiv:2407.09230
(2024).
arXiv:2407.09230 [cs.CV]
\end{botherref}
\endbibitem

%%% 17
\bibitem[\protect\citeauthoryear{Li et~al.}{2024}]{li2024endora}
\begin{barticle}
\bauthor{\bsnm{Li}, \binits{C.}},
\bauthor{\bsnm{Liu}, \binits{H.}},
\bauthor{\bsnm{Liu}, \binits{Y.}},
\bauthor{\bsnm{Feng}, \binits{B.Y.}},
\bauthor{\bsnm{Li}, \binits{W.}},
\bauthor{\bsnm{Liu}, \binits{X.}},
\bauthor{\bsnm{Chen}, \binits{Z.}},
\bauthor{\bsnm{Shao}, \binits{J.}},
\bauthor{\bsnm{Yuan}, \binits{Y.}}:
\batitle{Endora: Video generation models as endoscopy simulators}.
\bjtitle{arXiv preprint arXiv:2403.11050}
(\byear{2024})
\doiurl{10.48550/arXiv.2403.11050} .
\bcomment{Project page: this https URL}
\end{barticle}
\endbibitem

%%% 18
\bibitem[\protect\citeauthoryear{Iliash et~al.}{2024}]{iliash2024interactive}
\begin{botherref}
\oauthor{\bsnm{Iliash}, \binits{I.}},
\oauthor{\bsnm{Allmendinger}, \binits{S.}},
\oauthor{\bsnm{Meissen}, \binits{F.}},
\oauthor{\bsnm{K{\"u}hl}, \binits{N.}},
\oauthor{\bsnm{R{\"u}ckert}, \binits{D.}}:
Interactive generation of laparoscopic videos with diffusion models.
arXiv preprint arXiv:2406.06537
(2024).
arXiv:2406.06537 [eess.IV]
\end{botherref}
\endbibitem

%%% 19
\bibitem[\protect\citeauthoryear{Yang et~al.}{2024}]{yang2024cogvideox}
\begin{botherref}
\oauthor{\bsnm{Yang}, \binits{Z.}},
\oauthor{\bsnm{Teng}, \binits{J.}},
\oauthor{\bsnm{Zheng}, \binits{W.}},
\oauthor{\bsnm{Ding}, \binits{M.}},
\oauthor{\bsnm{Huang}, \binits{S.}},
\oauthor{\bsnm{Xu}, \binits{J.}},
\oauthor{\bsnm{Yang}, \binits{Y.}},
\oauthor{\bsnm{Hong}, \binits{W.}},
\oauthor{\bsnm{Zhang}, \binits{X.}},
\oauthor{\bsnm{Feng}, \binits{G.}}, et al.:
Cogvideox: Text-to-video diffusion models with an expert transformer.
arXiv preprint arXiv:2408.06072
(2024)
\end{botherref}
\endbibitem

%%% 20
\bibitem[\protect\citeauthoryear{Heusel et~al.}{2017}]{heusel2017ttur}
\begin{bchapter}
\bauthor{\bsnm{Heusel}, \binits{M.}},
\bauthor{\bsnm{Ramsauer}, \binits{H.}},
\bauthor{\bsnm{Unterthiner}, \binits{T.}},
\bauthor{\bsnm{Nessler}, \binits{B.}},
\bauthor{\bsnm{Hochreiter}, \binits{S.}}:
\bctitle{Gans trained by a two time-scale update rule converge to a local nash equilibrium}.
In: \bbtitle{Advances in Neural Information Processing Systems},
vol. \bseriesno{30}
(\byear{2017}).
\doiurl{10.48550/arXiv.1706.08500} .
\bcomment{arXiv:1706.08500}.
\burl{https://arxiv.org/abs/1706.08500}
\end{bchapter}
\endbibitem

%%% 21
\bibitem[\protect\citeauthoryear{Unterthiner et~al.}{2018}]{Unterthiner2018}
\begin{botherref}
\oauthor{\bsnm{Unterthiner}, \binits{T.}},
\oauthor{\bsnm{Steenkiste}, \binits{S.}},
\oauthor{\bsnm{Kurach}, \binits{K.}},
\oauthor{\bsnm{Marinier}, \binits{R.}},
\oauthor{\bsnm{Michalski}, \binits{M.}},
\oauthor{\bsnm{Gelly}, \binits{S.}}:
Towards accurate generative models of video: A new metric \& challenges
(2018)
{\href{https://arxiv.org/abs/1812.01717}{{arXiv:1812.01717}}}
{[cs.CV]}
\end{botherref}
\endbibitem

%%% 22
\bibitem[\protect\citeauthoryear{Rombach et~al.}{2021}]{LDM}
\begin{botherref}
\oauthor{\bsnm{Rombach}, \binits{R.}},
\oauthor{\bsnm{Blattmann}, \binits{A.}},
\oauthor{\bsnm{Lorenz}, \binits{D.}},
\oauthor{\bsnm{Esser}, \binits{P.}},
\oauthor{\bsnm{Ommer}, \binits{B.}}:
High-resolution image synthesis with latent diffusion models
(2021)
{\href{https://arxiv.org/abs/2112.10752}{{arXiv:2112.10752}}}
{[cs.CV]}
\end{botherref}
\endbibitem

%%% 23
\bibitem[\protect\citeauthoryear{Blattmann et~al.}{2023}]{ALIGN}
\begin{botherref}
\oauthor{\bsnm{Blattmann}, \binits{A.}},
\oauthor{\bsnm{Rombach}, \binits{R.}},
\oauthor{\bsnm{Ling}, \binits{H.}},
\oauthor{\bsnm{Dockhorn}, \binits{T.}},
\oauthor{\bsnm{Kim}, \binits{S.W.}},
\oauthor{\bsnm{Fidler}, \binits{S.}},
\oauthor{\bsnm{Kreis}, \binits{K.}}:
Align your latents: High-resolution video synthesis with latent diffusion models
(2023)
{\href{https://arxiv.org/abs/2304.08818}{{arXiv:2304.08818}}}
{[cs.CV]}
\end{botherref}
\endbibitem

%%% 24
\bibitem[\protect\citeauthoryear{Peebles and Xie}{2022}]{peebles2022scalable}
\begin{botherref}
\oauthor{\bsnm{Peebles}, \binits{W.}},
\oauthor{\bsnm{Xie}, \binits{S.}}:
Scalable diffusion models with transformers.
arXiv preprint
\textbf{2212.09748}
(2022)
{\href{https://arxiv.org/abs/2212.09748v2}{{arXiv:2212.09748v2}}}
{[cs.CV]}.
arXiv:2212.09748 [cs.CV], \url{https://doi.org/10.48550/arXiv.2212.09748}
\end{botherref}
\endbibitem

%%% 25
\bibitem[\protect\citeauthoryear{Lu et~al.}{2023}]{Lu2023VDT}
\begin{botherref}
\oauthor{\bsnm{Lu}, \binits{H.}},
\oauthor{\bsnm{Yang}, \binits{G.}},
\oauthor{\bsnm{Fei}, \binits{N.}},
\oauthor{\bsnm{Huo}, \binits{Y.}},
\oauthor{\bsnm{Lu}, \binits{Z.}},
\oauthor{\bsnm{Luo}, \binits{P.}},
\oauthor{\bsnm{Ding}, \binits{M.}}:
Vdt: General-purpose video diffusion transformers via mask modeling.
arXiv preprint arXiv:2305.13311
(2023).
Project page: https://VDT-2023.github.io
\end{botherref}
\endbibitem

%%% 26
\bibitem[\protect\citeauthoryear{Esser et~al.}{2024}]{esser2024scaling}
\begin{botherref}
\oauthor{\bsnm{Esser}, \binits{P.}},
\oauthor{\bsnm{Kulal}, \binits{S.}},
\oauthor{\bsnm{Blattmann}, \binits{A.}},
\oauthor{\bsnm{Entezari}, \binits{R.}},
\oauthor{\bsnm{Müller}, \binits{J.}},
\oauthor{\bsnm{Saini}, \binits{H.}},
\oauthor{\bsnm{Levi}, \binits{Y.}},
\oauthor{\bsnm{Lorenz}, \binits{D.}},
\oauthor{\bsnm{Sauer}, \binits{A.}},
\oauthor{\bsnm{Boesel}, \binits{F.}},
\oauthor{\bsnm{Podell}, \binits{D.}},
\oauthor{\bsnm{Dockhorn}, \binits{T.}},
\oauthor{\bsnm{English}, \binits{Z.}},
\oauthor{\bsnm{Lacey}, \binits{K.}},
\oauthor{\bsnm{Goodwin}, \binits{A.}},
\oauthor{\bsnm{Marek}, \binits{Y.}},
\oauthor{\bsnm{Rombach}, \binits{R.}}:
Scaling rectified flow transformers for high-resolution image synthesis.
arXiv preprint
\textbf{2403.03206}
(2024)
{\href{https://arxiv.org/abs/2403.03206v1}{{arXiv:2403.03206v1}}}
{[cs.CV]}.
arXiv:2403.03206 [cs.CV], \url{https://doi.org/10.48550/arXiv.2403.03206}
\end{botherref}
\endbibitem

%%% 27
\bibitem[\protect\citeauthoryear{Twinanda et~al.}{2016}]{twinanda2016endonet}
\begin{botherref}
\oauthor{\bsnm{Twinanda}, \binits{A.P.}},
\oauthor{\bsnm{Shehata}, \binits{S.}},
\oauthor{\bsnm{Mutter}, \binits{D.}},
\oauthor{\bsnm{Marescaux}, \binits{J.}},
\oauthor{\bsnm{Mathelin}, \binits{M.}},
\oauthor{\bsnm{Padoy}, \binits{N.}}:
Endonet: A deep architecture for recognition tasks on laparoscopic videos.
arXiv preprint
\textbf{1602.03012}
(2016)
{\href{https://arxiv.org/abs/1602.03012v2}{{arXiv:1602.03012v2}}}
{[cs.CV]}.
arXiv:1602.03012 [cs.CV], \url{https://doi.org/10.48550/arXiv.1602.03012}
\end{botherref}
\endbibitem

%%% 28
\bibitem[\protect\citeauthoryear{Kingma and Welling}{2013}]{kingma2013auto}
\begin{barticle}
\bauthor{\bsnm{Kingma}, \binits{D.P.}},
\bauthor{\bsnm{Welling}, \binits{M.}}:
\batitle{Auto-encoding variational bayes}.
\bjtitle{arXiv preprint arXiv:1312.6114}
(\byear{2013})
\doiurl{10.48550/arXiv.1312.6114} .
\bcomment{Fixes a typo in the abstract, no other changes}
\end{barticle}
\endbibitem

%%% 29
\bibitem[\protect\citeauthoryear{Yu et~al.}{2023}]{yu2023language}
\begin{barticle}
\bauthor{\bsnm{Yu}, \binits{L.}},
\bauthor{\bsnm{Lezama}, \binits{J.}},
\bauthor{\bsnm{Gundavarapu}, \binits{N.B.}},
\bauthor{\bsnm{Versari}, \binits{L.}},
\bauthor{\bsnm{Sohn}, \binits{K.}},
\bauthor{\bsnm{Minnen}, \binits{D.}},
\bauthor{\bsnm{Cheng}, \binits{Y.}},
\bauthor{\bsnm{Birodkar}, \binits{V.}},
\bauthor{\bsnm{Gupta}, \binits{A.}},
\bauthor{\bsnm{Gu}, \binits{X.}},
\bauthor{\bsnm{Hauptmann}, \binits{A.G.}},
\bauthor{\bsnm{Gong}, \binits{B.}},
\bauthor{\bsnm{Yang}, \binits{M.-H.}},
\bauthor{\bsnm{Essa}, \binits{I.}},
\bauthor{\bsnm{Ross}, \binits{D.A.}},
\bauthor{\bsnm{Jiang}, \binits{L.}}:
\batitle{Language model beats diffusion -- tokenizer is key to visual generation}.
\bjtitle{arXiv preprint arXiv:2310.05737}
(\byear{2023})
\doiurl{10.48550/arXiv.2310.05737} .
\bcomment{ICLR 2024}
\end{barticle}
\endbibitem

%%% 30
\bibitem[\protect\citeauthoryear{Arnab et~al.}{2021}]{arnab2021vivit}
\begin{bchapter}
\bauthor{\bsnm{Arnab}, \binits{A.}},
\bauthor{\bsnm{Dehghani}, \binits{M.}},
\bauthor{\bsnm{Heigold}, \binits{G.}},
\bauthor{\bsnm{Sun}, \binits{C.}},
\bauthor{\bsnm{Lu{\v{c}}i{\'c}}, \binits{M.}},
\bauthor{\bsnm{Schmid}, \binits{C.}}:
\bctitle{Vivit: A video vision transformer}.
In: \bbtitle{Proceedings of the IEEE/CVF International Conference on Computer Vision (ICCV)}
(\byear{2021}).
\doiurl{10.48550/arXiv.2103.15691} .
\bcomment{ICCV 2021. Code available at \url{https://github.com/yourlink}}.
\burl{https://arxiv.org/abs/2103.15691}
\end{bchapter}
\endbibitem

%%% 31
\bibitem[\protect\citeauthoryear{Raffel et~al.}{2019}]{raffel2019exploring}
\begin{barticle}
\bauthor{\bsnm{Raffel}, \binits{C.}},
\bauthor{\bsnm{Shazeer}, \binits{N.}},
\bauthor{\bsnm{Roberts}, \binits{A.}},
\bauthor{\bsnm{Lee}, \binits{K.}},
\bauthor{\bsnm{Narang}, \binits{S.}},
\bauthor{\bsnm{Matena}, \binits{M.}},
\bauthor{\bsnm{Zhou}, \binits{Y.}},
\bauthor{\bsnm{Li}, \binits{W.}},
\bauthor{\bsnm{Liu}, \binits{P.J.}}:
\batitle{Exploring the limits of transfer learning with a unified text-to-text transformer}.
\bjtitle{arXiv preprint arXiv:1910.10683}
(\byear{2019})
\doiurl{10.48550/arXiv.1910.10683} .
\bcomment{Version 4, updated on 19 Sep 2023}
\end{barticle}
\endbibitem

%%% 32
\bibitem[\protect\citeauthoryear{Chambon et~al.}{2022}]{Chambon2022AdaptingPV}
\begin{botherref}
\oauthor{\bsnm{Chambon}, \binits{P.}},
\oauthor{\bsnm{Bluethgen}, \binits{C.}},
\oauthor{\bsnm{Langlotz}, \binits{C.P.}},
\oauthor{\bsnm{Chaudhari}, \binits{A.}}:
Adapting pretrained vision-language foundational models to medical imaging domains.
Foundation Models for Decision Making Workshop at Neural Information Processing Systems
\textbf{arXiv:2210.04133}
(2022)
\end{botherref}
\endbibitem

%%% 33
\bibitem[\protect\citeauthoryear{Abaid et~al.}{2024}]{abaid2024synthesizing}
\begin{botherref}
\oauthor{\bsnm{Abaid}, \binits{A.}},
\oauthor{\bsnm{Farooq}, \binits{M.A.}},
\oauthor{\bsnm{Hynes}, \binits{N.}},
\oauthor{\bsnm{Corcoran}, \binits{P.}},
\oauthor{\bsnm{Ullah}, \binits{I.}}:
Synthesizing cta image data for type-b aortic dissection using stable diffusion models.
arXiv preprint arXiv:2402.06969
(2024)
{\href{https://arxiv.org/abs/2402.06969}{{arXiv:2402.06969}}}
{[cs.CV]}.
Submitted in IEEE EMBC 2024 Conference
\end{botherref}
\endbibitem

%%% 34
\bibitem[\protect\citeauthoryear{Loshchilov and Hutter}{2017}]{loshchilov2017decoupled}
\begin{botherref}
\oauthor{\bsnm{Loshchilov}, \binits{I.}},
\oauthor{\bsnm{Hutter}, \binits{F.}}:
Decoupled weight decay regularization.
arXiv preprint arXiv:1711.05101
(2017)
\end{botherref}
\endbibitem

%%% 35
\bibitem[\protect\citeauthoryear{He et~al.}{2016}]{he2016deep}
\begin{bchapter}
\bauthor{\bsnm{He}, \binits{K.}},
\bauthor{\bsnm{Zhang}, \binits{X.}},
\bauthor{\bsnm{Ren}, \binits{S.}},
\bauthor{\bsnm{Sun}, \binits{J.}}:
\bctitle{Deep residual learning for image recognition}.
In: \bbtitle{Proceedings of the IEEE Conference on Computer Vision and Pattern Recognition},
pp. \bfpage{770}--\blpage{778}
(\byear{2016})
\end{bchapter}
\endbibitem

%%% 36
\bibitem[\protect\citeauthoryear{Hara et~al.}{2017}]{hara2017learning}
\begin{barticle}
\bauthor{\bsnm{Hara}, \binits{K.}},
\bauthor{\bsnm{Kataoka}, \binits{H.}},
\bauthor{\bsnm{Satoh}, \binits{Y.}}:
\batitle{Learning spatio-temporal features with 3d residual networks for action recognition}.
\bjtitle{arXiv preprint arXiv:1708.07632}
(\byear{2017})
\doiurl{10.48550/arXiv.1708.07632} .
\bcomment{To appear in ICCV 2017 Workshop (Chalearn)}
\end{barticle}
\endbibitem

%%% 37
\bibitem[\protect\citeauthoryear{Kay et~al.}{2017}]{kay2017kinetics}
\begin{botherref}
\oauthor{\bsnm{Kay}, \binits{W.}},
\oauthor{\bsnm{Carreira}, \binits{J.}},
\oauthor{\bsnm{Simonyan}, \binits{K.}},
\oauthor{\bsnm{Zhang}, \binits{B.}},
\oauthor{\bsnm{Hillier}, \binits{C.}},
\oauthor{\bsnm{Vijayanarasimhan}, \binits{S.}},
\oauthor{\bsnm{Viola}, \binits{F.}},
\oauthor{\bsnm{Green}, \binits{T.}},
\oauthor{\bsnm{Back}, \binits{T.}},
\oauthor{\bsnm{Natsev}, \binits{P.}}, et al.:
The kinetics human action video dataset.
arXiv preprint arXiv:1705.06950
(2017)
\end{botherref}
\endbibitem

%%% 38
\bibitem[\protect\citeauthoryear{Kim et~al.}{2021}]{kim2021drivegan}
\begin{bchapter}
\bauthor{\bsnm{Kim}, \binits{S.W.}},
\bauthor{\bsnm{Philion}, \binits{J.}},
\bauthor{\bsnm{Torralba}, \binits{A.}},
\bauthor{\bsnm{Fidler}, \binits{S.}}:
\bctitle{Drivegan: Towards a controllable high-quality neural simulation}.
In: \bbtitle{Proceedings of the IEEE/CVF Conference on Computer Vision and Pattern Recognition (CVPR)}
(\byear{2021}).
\doiurl{10.48550/arXiv.2104.15060} .
\bcomment{Oral}.
\burl{https://doi.org/10.48550/arXiv.2104.15060}
\end{bchapter}
\endbibitem

%%% 39
\bibitem[\protect\citeauthoryear{Hu et~al.}{2023}]{hu2023gaia}
\begin{botherref}
\oauthor{\bsnm{Hu}, \binits{A.}},
\oauthor{\bsnm{Russell}, \binits{L.}},
\oauthor{\bsnm{Yeo}, \binits{H.}},
\oauthor{\bsnm{Murez}, \binits{Z.}},
\oauthor{\bsnm{Fedoseev}, \binits{G.}},
\oauthor{\bsnm{Kendall}, \binits{A.}},
\oauthor{\bsnm{Shotton}, \binits{J.}},
\oauthor{\bsnm{Corrado}, \binits{G.}}:
Gaia-1: A generative world model for autonomous driving.
arXiv preprint arXiv:2309.17080
(2023)
\end{botherref}
\endbibitem

%%% 40
\bibitem[\protect\citeauthoryear{Wang et~al.}{2023}]{wang2023drivedreamer}
\begin{botherref}
\oauthor{\bsnm{Wang}, \binits{X.}},
\oauthor{\bsnm{Zhu}, \binits{Z.}},
\oauthor{\bsnm{Huang}, \binits{G.}},
\oauthor{\bsnm{Chen}, \binits{X.}},
\oauthor{\bsnm{Lu}, \binits{J.}}:
Drivedreamer: Towards real-world-driven world models for autonomous driving.
arXiv preprint arXiv:2309.09777
(2023)
\end{botherref}
\endbibitem

%%% 41
\bibitem[\protect\citeauthoryear{Badash et~al.}{2016}]{Badash2016}
\begin{barticle}
\bauthor{\bsnm{Badash}, \binits{I.}},
\bauthor{\bsnm{Burtt}, \binits{K.}},
\bauthor{\bsnm{Solorzano}, \binits{C.A.}},
\bauthor{\bsnm{Carey}, \binits{J.N.}}:
\batitle{Innovations in surgery simulation: a review of past, current and future techniques}.
\bjtitle{Annals of Translational Medicine}
\bvolume{4}(\bissue{23}),
\bfpage{453}
(\byear{2016})
\doiurl{10.21037/atm.2016.12.24}
\end{barticle}
\endbibitem

\end{thebibliography}

\end{document}